\documentclass[11pt]{article}
\usepackage{eacl2017}
\usepackage{times}
\usepackage{url}
\usepackage{latexsym}

\eaclfinalcopy

\usepackage[utf8]{inputenc}
\usepackage{graphicx}
\usepackage{url}
\usepackage{multirow}
\usepackage{todonotes}
\usepackage{xcolor}

\definecolor{blue}{RGB}{0,0,0}
\definecolor{purple}{RGB}{0,0,0}
\definecolor{green}{RGB}{0,0,0}
\definecolor{red}{RGB}{0,0,0}
\definecolor{orange}{RGB}{0,0,0}

\title{A Multifaceted Evaluation of Neural versus Phrase-Based Machine Translation for 9 Language Directions}

\author{Antonio Toral\thanks{~~Work partly done at his previous position in Dublin City University, Ireland.}\\
	University of Groningen\\
    The Netherlands\\
  {\tt a.toral.ruiz@rug.nl} \\\And
  V{\'i}ctor M. S{\'a}nchez-Cartagena\\
	    Prompsit Language Engineering \\
        Av. Universitat s/n. Edifici Quorum III\\
	    E-03202 Elx, Spain\\
	    {\tt vmsanchez@prompsit.com}}

\date{}

\begin{document}
\maketitle
\begin{abstract}
We aim to shed light on the strengths and weaknesses of the newly introduced neural machine translation paradigm.
To that end, we conduct a multifaceted evaluation in which we compare outputs produced by state-of-the-art neural machine translation and phrase-based machine translation systems for 9 language directions across a number of dimensions.
Specifically, we measure the similarity of the outputs, their fluency and amount of reordering, the effect of sentence length and performance across different error categories. 
We find out that translations produced by neural machine translation systems are 
considerably different, more fluent and more accurate in terms of word order 
compared to those produced by phrase-based systems.
Neural machine translation systems
are also more accurate at producing inflected forms, but they perform poorly when translating very long sentences.
\end{abstract}

\section{Introduction}

A new paradigm to statistical machine translation, neural MT (NMT), has emerged very recently
and has already surpassed the performance of the mainstream approach in the field, phrase-based MT (PBMT), for a number of language pairs, e.g.~\cite{sennrich2015b,luong15,DBLP:journals/corr/Costa-JussaF16,DBLP:journals/corr/ChungCB16}.

In PBMT~\cite{Koehn:2010:SMT:1734086} different models (translation, reordering, target language, etc.) are trained independently and combined in a log-linear scheme in which each model is assigned a different weight by a tuning algorithm. On the contrary, in NMT all the components are jointly trained to maximise translation quality. NMT systems have a strong generalisation power because they encode translation units as numeric vectors that represent concepts, whereas in PBMT translation units are encoded as strings. Moreover, NMT systems are able to model long-distance phenomena thanks to the use of recurrent neural networks, e.g. long short-term memory (LSTM)~\cite{hochreiter1997long} or gated recurrent units~\cite{DBLP:journals/corr/ChungGCB14}.

The translations produced by NMT systems have been evaluated thus far mostly in terms of overall performance scores, be it by means of automatic or human evaluations.
This has been the case of last year's news translation shared task at the First Conference on Machine Translation (WMT16).\footnote{\url{http://www.statmt.org/wmt16/translation-task.html}}
In this translation task, outputs produced by participant MT systems, the vast majority of which fall under either the \textcolor{purple}{phrase-based} or neural approaches, were evaluated (i) automatically with the BLEU~\cite{papineni2002bleu} and TER~\cite{snover2006study} metrics, and (ii) manually by means of ranking translations~\cite{mtm12_appraise} and monolingual semantic similarity~\cite{NLE:9961497}.
In all these evaluations, the performance of each system is measured by means of an overall score, which, while giving an indication of the general performance of a given system, does not provide any additional information.

In order to understand better the new NMT paradigm and in what respects it provides better (or worse) translation quality than state-of-the-art PBMT, Bentivogli et al.~\shortcite{Bentivogli1608.04631} conducted a detailed analysis for the English-to-German language direction.
In a nutshell, they found out that NMT (i) decreases post-editing effort, (ii) degrades faster than PBMT with sentence length and (iii) results in a notable improvement regarding reordering.

In this paper we delve further in this direction by conducting a multilingual and multifaceted evaluation in order to find answers to the following research questions. Whether, in comparison to PBMT, NMT systems result in:
\begin{itemize}
\item 
\textcolor{purple}{considerably} different output \textcolor{blue}{and higher degree of variability}; 
\item more or less fluent output; 
\item more or less monotone translations; 
\item \textcolor{red}{translations with better or worse word order;}
\item better or worse translations depending on sentence length;
\item less or more errors for different error categories: inflectional, reordering and lexical;
\end{itemize}

Hereunder we specify the main differences and similarities between this work and that of Bentivogli et al.~\shortcite{Bentivogli1608.04631}:
\begin{itemize}
\item Language directions. They considered 1 while our study comprises 9.
\item Content. They dealt with transcribed speeches while we work with news stories. Previous research has shown that these two types of content pose different challenges for MT~\cite{ruiz2014complexity}.
\item Size of evaluation data. Their test set had 600 sentences while our test sets span from $1\,999$ to $3\,000$ depending on the language direction.
\item Reference type. Their references were both independent from the MT output and also post-edited, while we have access only to single independent references.
\item Analyses. While some analyses overlap, some are novel in our experiments. Namely, output \textcolor{purple}{similarity}, fluency and degree of \textcolor{purple}{reordering performed}. 
\end{itemize}

Our analyses are conducted on the best PBMT and NMT systems submitted to the WMT16 translation task for each language direction.
This (i) guarantees the reproducibility of our results as all the MT outputs are publicly available,
(ii) ensures that the systems evaluated are state-of-the-art, as they are the result of the latest developments at top MT research groups worldwide, and 
(iii) allows the conclusions that will be drawn to be rather general, as 6 languages from 4 different families (Germanic, Slavic, Romance and Finno-Ugric) are covered in the experiments.

The rest of the paper is organised as follows.
Section \ref{s_experimental_setup} describes the experimental setup.
Subsequent sections cover the experiments carried out in which we measured different aspects of NMT, namely: \textcolor{purple}{output similarity} (Section~\ref{s_overlap}), fluency (Section~\ref{s_fluency}), 
\textcolor{red}{degree of reordering and quality of word order}
(Section~\ref{s_monotonicity}), sentence length (Section~\ref{s_sentence_length}), and amount of errors for different error categories (Section~\ref{s_error_types}). 
Finally, Section \ref{s_conclusions} holds the conclusions and proposals for future work.

\section{Experimental Setup}\label{s_experimental_setup}

The experiments are run on the best\footnote{
According to the human evaluation~\cite[Sec. 3.4]{bojar-EtAl:2016:WMT1}. When there are not statistically significant differences between two or more NMT or PBMT systems (i.e. they belong to the same equivalence class), we pick the one with the highest BLEU score. If two NMT or PBMT systems were the best according to BLEU (draw), we pick the one with the best TER score.
} 
PBMT\footnote{Many of the PBMT systems contain neural features, mainly in the form of language models. 
If the best PBMT submission contains any neural features we use this as the PBMT system in our analyses as long as none of these features is a fully-fledged NMT system. This was the case of the best submission in terms of BLEU for RU$\rightarrow$EN~\cite{junczysdowmunt-dwojak-sennrich:2016:WMT}
}
and NMT constrained systems submitted to the news translation task of WMT16.
Out of the 12 language directions at the translation task, we conduct experiments on 9.\footnote{Some experiments are run on a subset of these languages due to the lack of required tools for some of the languages involved.}
These are the language pairs between English (EN) and Czech (CS), German (DE), Finnish (FI), Romanian (RO) and Russian (RU) in both directions (except for Finnish, where only the EN$\rightarrow$FI direction is covered as no NMT system was submitted for the opposite direction, FI$\rightarrow$EN).
Finally, there was an additional language at the shared task, Turkish, that is not considered here, as either none of the systems submitted was neural (Turkish$\rightarrow$EN), or there was one such system but its performance was extremely low (EN$\rightarrow$Turkish)
and hence most probably not representative of the state-of-the-art in NMT.

\begin{small}
\begin{table*}
\centering
\begin{tabular}{ p{1.5cm} | p{1.7cm} | p{11.3cm} }
\bf Language Pair  & \bf MT Paradigm &\bf System details\\
\hline
\multirow{2}{*}{EN$\rightarrow$CS}	&PBMT	&Phrase-based, word clusters~\cite{ding-EtAl:2016:WMT}\\
					&NMT	&Unsupervised word segmentation and backtranslated monolingual corpora~\cite{sennrich-haddow-birch:2016:WMT}\\
\hline
\multirow{2}{*}{EN$\rightarrow$DE}	&hierarchical PBMT	&String-to-tree, neural and dependency language models~\cite{williams-EtAl:2016:WMT}\\
					&NMT	&Same as for EN$\rightarrow$CS\\
\hline
\multirow{2}{*}{EN$\rightarrow$FI}	&PBMT	&Phrase-based, rule-based and unsupervised word segmentation, operation sequence model~\cite{durrani2011joint}, bilingual neural language model~\cite{devlin-EtAl:2014:P14-1}, re-ranked with a recurrent neural language model~\cite{sanchezcartagena-toral:2016:WMT}\\
					&NMT	&Rule-based word segmentation, backtranslated monolingual corpora~\cite{sanchezcartagena-toral:2016:WMT}\\
\hline
\multirow{2}{*}{EN$\rightarrow$RO}	&PBMT	&Phrased-based, operation sequence model, monolingual and bilingual neural language models~\cite{williams-EtAl:2016:WMT}\\
					&NMT	&Same as for EN$\rightarrow$CS\\
\hline
\multirow{2}{*}{EN$\rightarrow$RU}	&PBMT	&Phrase-based, word clusters, bilingual neural language model~\cite{ding-EtAl:2016:WMT}\\
					&NMT	&Same as for EN$\rightarrow$CS\\
\hline
\multirow{2}{*}{CS$\rightarrow$EN}	&PBMT	&Same as for EN$\rightarrow$CS\\
					&NMT	&Same as for EN$\rightarrow$CS\\
\hline
\multirow{2}{*}{DE$\rightarrow$EN}	&PBMT	&Phrase-based, pre-reordering, compound splitting~\cite{williams-EtAl:2016:WMT}\\
					&NMT	&Same as for EN$\rightarrow$CS plus reranked with a right-to-left model\\
\hline
\multirow{2}{*}{RO$\rightarrow$EN}	&PBMT	&Phrase-based, operation sequence model, monolingual neural language model~\cite{williams-EtAl:2016:WMT}\\
					&NMT	&Same as for EN$\rightarrow$CS\\
\hline
\multirow{2}{*}{RU$\rightarrow$EN}	&PBMT	&Phrase-based, lemmas in word alignment, sparse features, bilingual neural language model and transliteration~\cite{lo-EtAl:2016:WMT}\\
					&NMT	&Same as for EN$\rightarrow$CS\\
\hline
\end{tabular}
\caption{Details of the best systems pertaining to the PBMT and NMT paradigms submitted to the WMT16 news translation task 
for each language direction.}
\label{t:smt_nmt_system_details}
\end{table*}
\end{small}

Table \ref{t:smt_nmt_system_details} shows the main characteristics of the best PBMT and NMT systems submitted to the WMT16 news translation task. 
It should be noted that all \textcolor{purple}{the} NMT systems listed in the table fall under the encoder-decoder architecture with attention~\cite{Bahdanau2014} and operate on subword units. Word segmentation is carried out with the help of a lexicon in the EN$\rightarrow$FI direction~\cite{sanchezcartagena-toral:2016:WMT} and in an unsupervised way in the \textcolor{purple}{remaining} directions~\cite{sennrich-haddow-birch:2016:WMT}.

\subsection{Overall Evaluation}

First, and in order to contextualise our analyses below,
we report the BLEU scores achieved by the best NMT and PBMT systems
for each language direction at WMT16's news translation task in Table \ref{t:smt_nmt_overall_results}.\footnote{We report the official results from \url{http://matrix.statmt.org/matrix} for the test set \emph{newstest2016} using normalised BLEU (column \emph{z BLEU-cased-norm}).}
The best NMT system clearly outperforms the best PBMT system for all language directions out of English (relative improvements range from 5.5\% for EN$\rightarrow$RO to 17.6\% for EN$\rightarrow$FI) and \textcolor{blue}{the} human evaluation~\cite[Sec. 3.4]{bojar-EtAl:2016:WMT1} confirms these results. 
In the opposite direction, the human evaluation shows that the best NMT system outperforms the best PBMT system for all language directions except when the source language is Russian. This slightly differs from the automatic evaluation, according to which NMT outperforms PBMT for translations from Czech (3.3\% relative improvement) and German (9.9\%) but underperforms PBMT for translations from Romanian (-3.7\%) and Russian (-3.8\%).

\begin{table}[htbp]
\centering
\begin{tabular}{ l | r r r r r}
\bf System  & \bf CS	& \bf DE	& \bf FI	&\bf RO	& \bf RU\\
\hline
	&\multicolumn{5}{c}{From EN}\\
PBMT	&$23.7$	&$30.6$	&$15.3$	&$27.4$	&$24.3$\\ 
NMT	&$\mathbf{25.9}$	&$\mathbf{34.2}$	&$\mathbf{18.0}$	&$\mathbf{28.9}$	&$\mathbf{26.0}$\\
\hline
	&\multicolumn{5}{c}{Into EN}\\
PBMT	&$30.4$	&$35.2$	&$23.7$	&$\mathbf{35.4}$	&$\mathbf{29.3}$\\
NMT	&$\mathbf{31.4}$	&$\mathbf{38.7}$	&-	&$34.1$	&$28.2$\\
\hline
\end{tabular}
\caption{BLEU scores of the best NMT and PBMT systems for each language pair at WMT16's news translation task. If the difference between them is statistically significant according to paired bootstrap resampling~\cite{koehn2004statistical} with $p = 0.05$ and $1\,000$ iterations, the highest score is shown in bold.
}
\label{t:smt_nmt_overall_results}
\end{table}


%
%
%
%
%
%
%
%

\section{Output \textcolor{purple}{Similarity}}\label{s_overlap}

The aim of this analysis is to \textcolor{purple}{assess to which extent} translations produced by NMT systems are different from those produced by PBMT systems.
We measure this by taking the outputs of the top $n$\footnote{The number of systems considered is different for each language direction as it depends on the number of systems submitted. Namely, we have considered 2 NMT and 2 PBMT into Czech, 3 NMT and 5 PBMT into German, 2 NMT and 4 PBMT into Finnish, 2 NMT and 4 PBMT into Romanian and 2 NMT and 3 PBMT into Russian.} NMT and PBMT systems \textcolor{blue}{submitted} to each language direction\footnote{In order to make sure that all systems considered are truly different (rather than different runs of the same system) we consider only 1 system \textcolor{blue}{per paradigm (NMT and PBMT) submitted by each team for each language direction.}} and checking their pairwise overlap in terms of the chrF1~\cite{popovic:2015:WMT} automatic evaluation metric.\footnote{Throughout our analyses we use this metric as it has been shown to correlate better with human judgements than the {\it de facto} standard automatic metric, BLEU, when the target language is a morphologically rich language such as Finnish, while its correlation is on par with BLEU for languages with simpler morphology such as English~\cite{popovic:2015:WMT}.}

We would consider NMT outputs \textcolor{purple}{considerably different (with respect to PBMT)} if they resemble each other (i.e. high pairwise overlap between NMT outputs) more than they do to PBMT systems (i.e. low overlap between an output by NMT and another by PBMT).
This analysis is carried out only for language directions out of English, as for all the language directions into English there was, at most, 1 NMT submission.

\begin{table}[htbp]
\centering
\begin{tabular}{ l | r r r r}
\bf TL	&\bf 2 NMT &\bf 2 PBMT	&\bf NMT \& PBMT\\
\hline
 CS	&$68.66$	&$77.63$	&$64.34$\\
DE	&$72.10$	&$72.97$	&$66.80$\\
 FI	&$56.03$	&$57.42$	&$55.55$\\
 RO	&$69.47$	&$75.96$	&$68.77$\\
 RU	&$35.52$	&$43.35$	&$29.87$\\
\hline
\end{tabular}
\caption{Average of the overlaps between pairs of outputs produced by the top $n$ NMT and PBMT systems for each language direction from English to the target language (TL). The higher the value, the larger is the overlap.  } 
\label{t:overlap}
\end{table}

Table \ref{t:overlap} shows the results. We can observe the same trends for all the language directions, namely:
(i) the highest overlaps are between pairs of PBMT systems; 
(ii) next, we have overlaps between NMT systems;
(iii) finally, overlaps between PBMT and NMT are the lowest.

We can conclude then that NMT systems lead to \textcolor{purple}{considerably} different outputs compared to PBMT.
The fact that \textcolor{purple}{there is higher inter-system variability in NMT than in PBMT (i.e. overlaps between pairs of NMT systems are lower than between pairs of PBMT systems)} may surprise the reader, considering the fact that all NMT systems belong to the same paradigm (encoder-decoder with attention) while for some language directions (EN$\rightarrow$DE, EN$\rightarrow$FI and EN$\rightarrow$RO) there are PBMT systems belonging to two different paradigms (pure phrase-based and hierarchical).
However, the higher variability among NMT translations \textcolor{blue}{can be attributed, we believe,} to the fact that NMT systems use numeric vectors that represent concepts instead of strings as translation units.

\section{Fluency}\label{s_fluency}

In this experiment we aim to find out whether the outputs produced by NMT systems are more or less fluent than those produced by PBMT systems.
To that end, we take perplexity of the MT outputs on neural language models (LMs) as a proxy for fluency.
The LMs are built using \texttt{TheanoLM}~\cite{theanolm2016}. They contain $100$ units 
in the projection layer, $300$ units 
in the LSTM layer, and $300$ units 
in the \emph{tanh} layer, following the setup described by Enarvi and Kurimo~\shortcite[Sec. 3.2]{theanolm2016}. The training algorithm is Adagrad~\cite{duchi2011adaptive} and \textcolor{blue}{we used} $1\,000$ word classes obtained with \texttt{mkcls} from the training corpus. Vocabulary is limited to the most frequent $50\,000$ tokens.

LMs are trained on a random sample of 4 million sentences selected from the News Crawl 2015 monolingual corpora, available for all the languages considered.\footnote{\url{http://data.statmt.org/wmt16/translation-task/training-monolingual-news-crawl.tgz}}

\begin{table}[htbp]
\centering
\begin{tabular}{ l | r r r}
\bf Language 	& \multirow{2}{*}{\bf PBMT}	& \multirow{2}{*}{\bf NMT}	& \multirow{2}{*}{\bf Rel. diff.} \\
\bf direction	& 	& &   \\

\hline
EN$\rightarrow$CS	& $202.91$ & $173.33$ & $-14.58\%$ \\
EN$\rightarrow$DE	& $131.54$	& $107.08$ & $-18.60\%$ \\
EN$\rightarrow$FI	& $214.10$ & $222.40$ & $3.88\%$\\
EN$\rightarrow$RO	& $124.66$	& $116.33$	& $-6.68\%$ \\
EN$\rightarrow$RU	& $158.18$ & $127.83$ & $-19.19\%$ \\
\hline
CS$\rightarrow$EN	& $110.08$ & $102.36$ &$-7.01\%$\\
DE$\rightarrow$EN	& $122.26$& $104.72$ & $-14.35\%$\\
RO$\rightarrow$EN	& $106.08$	& $102.18$ & $-3.68\%$ \\
RU$\rightarrow$EN	& $123.86$ & $106.75$ & $-13.81\%$\\

\hline
Average			& $143.74$ & $129.22$ & $-10.45\%$\\
\hline
\end{tabular}
\caption{Perplexity scores for the outputs of the best NMT and PBMT systems on language models built on $4$ million sentences randomly selected from the News Crawl 2015 corpora.}
\label{t:fluency}
\end{table}

Table \ref{t:fluency} shows the results. For all the language directions considered but one, perplexity is higher on the PBMT output compared to the NMT output.
The only exception is translation into Finnish, in which
perplexity on the PBMT output is slightly lower, probably because its fluency was improved  by reranking it with a neural LM similar to the one we use in this experiment~\cite{sanchezcartagena-toral:2016:WMT}. 
The average relative difference, i.e. considering all language directions, is notable at $-10.45\%$. Thus, our experiment shows that the outputs produced by NMT systems are, in general, more fluent than those produced by PBMT systems.

One may argue that the perplexity obtained for NMT outputs is lower than that for PBMT outputs because the LMs we used to measure perplexity follow the same model as the decoder of the NMT architecture~\cite{Bahdanau2014} 
and hence perplexity on a neural LM is not a valid proxy for fluency. However, the following facts 
support our strategy:
\begin{itemize}
\item The manual evaluation of fluency carried out at the WMT16 shared translation task~\cite[Sec. 3.5]{bojar-EtAl:2016:WMT1} already confirmed that NMT systems consistently produce more fluent translations than PBMT systems. That manual evaluation only covered language directions into English. In this experiment, we extend that conclusion to 
language directions out of English.
\item Neural LMs consistently outperform $n$-gram based LMs when assessing the fluency of \emph{real} text~\cite{AAAI1612489,theanolm2016}. Thus, we have used the most accurate automatic tool available to measure fluency.
\end{itemize}

\begin{table*}
\centering
\begin{tabular}{ l | r r r | r r}
\multirow{2}{*}{\bf Language direction} & \multicolumn{3}{|c|}{\bf Monotone vs.}  & \multirow{2}{*}{\bf  PBMT vs. Ref.} &  \multirow{2}{*}{\bf  NMT vs. Ref.} \\
	&\bf PBMT	&\bf NMT &\bf Ref.	& & \\
\hline
EN$\rightarrow$CS	&$0.9273$	& $0.9029$ & $0.8295$ & $\mathbf{0.8008}$ & $0.7964$  \\
EN$\rightarrow$DE	&$0.8006$	&$0.8229$	& $0.7740$  & $0.7560$  & $\mathbf{0.7791}$ \\
EN$\rightarrow$FI	&$0.8912$	&$0.9172$	& $0.8367$  & $0.7611$  & $\mathbf{0.7819}$ \\
EN$\rightarrow$RO	&$0.8389$	&$0.8378$	& $0.7937$  & $0.8312$ & $0.8282$ \\
EN$\rightarrow$RU	&$0.9342$	& $0.9114$ & $0.8364$  & $0.8249$  & $0.8240$ \\
\hline
CS$\rightarrow$EN	& $0.7694$ & $0.7589$ & $0.7128$  & $0.8000$  & $0.8015$ \\
DE$\rightarrow$EN	&$0.8036$	&$0.7830$	& $0.7409$ & $0.7728$ & $\mathbf{0.7943}$ \\
RO$\rightarrow$EN	&$0.8693$	&$0.8427$	& $0.8013$ & $0.8187$ & $\mathbf{0.8245}$ \\
RU$\rightarrow$EN	& $0.8170$ & $0.7891$ & $0.7247$ & $0.7958$  & $\mathbf{0.8069}$ \\
\hline
\end{tabular}
\caption{ Average Kendall's tau distance between the word alignments obtained after translating the test set with each MT system being evaluated and a monotone alignment (left); and average Kendall's tau distance between the word alignments obtained 
\textcolor{purple}{for} each MT system's translation and the word alignments of the reference translation (right). Larger values represent more similar alignments. If the difference between the distances depicted in the two last columns is statistically significant according to paired bootstrap resampling~\cite{koehn2004statistical} with $p = 0.05$ and $1\,000$ iterations, the largest distance is shown in bold.}
\label{t:monotonicity}
\end{table*}

\section{Reordering}\label{s_monotonicity}

In this section we measure the amount of reordering performed by PBMT and NMT systems. 
Our objective is to empirically determine whether: (i) the recurrent neural networks in NMT systems produce more changes in the word order of a sentence than an PBMT decoder; and whether (ii) these 
neural networks make the word order of the translations closer to that of the reference.

In order to measure the amount of reordering, we used the \emph{Kendall's tau distance} between word alignments obtained from pairs of sentences~\cite[Sec. 5.3.2]{birch2011reordering}. As the distance needs to be computed from permutations,\footnote{A permutation between a source-language sentence and a target-language sentence is defined as the set of operations that need to be carried out over the words in the source-language sentence to reflect the order of the words in the target-language sentence~\cite[Sec. 5.2]{birch2011reordering}.} we turned word aligments into permutations by means of the algorithm defined by Birch~\shortcite[Sec. 5.2]{birch2011reordering}.

For each language direction, we computed word alignments between the source-language side of the test set and the target-language reference, the PBMT output and the NMT output by means of \texttt{MGIZA++}~\cite{Gao:2008:PIW:1622110.1622119}.
As the test sets are rather small for word alignment ($1\,999$ to $3\,000$ sentence pairs depending on the language pair), we append bigger parallel corpora to help ensure accurate word alignments and avoid data sparseness.
For languages for which in-domain (news) parallel training data is available (German and Russian), we append that dataset (News Commentary).
For the remaining languages (Finnish and Romanian) we use the whole Europarl corpus.

The amount of reordering performed by each system can be estimated as the distance between the word alignments produced by that system and a monotone word alignment. The similarity between the reorderings produced by each MT system and the reorderings in the reference translation can also be estimated as the distance between the corresponding word alignments. Table~\ref{t:monotonicity} shows the value of these distances for the language pairs included in our evaluation. The average over all the sentences in the test set of the distance proposed by Birch~\shortcite{birch2011reordering} is depicted.

It can be observed that the amount of reordering introduced by both types of MT systems is lower than the quantity of reordering in the reference translation. 
NMT generally produces more changes in the structure of the sentence than PBMT. This is the case for all language pairs but \textcolor{purple}{two (}EN$\rightarrow$DE and EN$\rightarrow$FI). A possible explanation for these two exceptions is the following: in the former language pair, the PBMT system is hierarchical~\cite{williams-EtAl:2016:WMT} while in the latter, the output was reranked with neural LMs.


Concerning the similarity between the reorderings produced by both MT systems and those in the reference translation,
out of 9 directions, in 5 directions the NMT system performs a reordering closer to the reference, in 1 direction the PBMT system performs a reordering closer to the reference and \textcolor{purple}{in the remaining 3 directions the differences are not statistically significant.}
That is, NMT generally produces reorderings which are closer to the reference translation.
The exceptions
to this trend, however, do not exactly correspond to the
language pairs for which NMT underperformed PBMT.

In summary, NMT systems achieve, in general, a higher degree of reordering than pure, phrase-based PBMT systems, and, 
\textcolor{purple}{overall, this reordering results in translations whose word order is closer to that of the reference translation.}

\section{Sentence Length}\label{s_sentence_length}

In this experiment we aim to find out whether the performances of NMT and PBMT are somehow sensitive to sentence length.
In this regard, Bentivogli et al.~\shortcite{Bentivogli1608.04631} found that, for transcribed speeches, NMT outperformed PBMT regardless of sentence length while also noted that NMT's performance degraded faster than PBMT's as sentence length increases.
It should be noted, however, that sentences in our content type, news, are considerably longer than sentences in transcribed speeches.\footnote{According to Ruiz et al.~\shortcite{ruiz2014complexity}, sentences of transcribed speeches in English average to 19 words while sentences in news average to 24 words.}
Hence, the current experiment will determine to what extent the findings on transcribed speeches stand also for texts made of longer sentences.

\begin{figure}[htbp]
 \centering
 \includegraphics[width=0.48\textwidth]{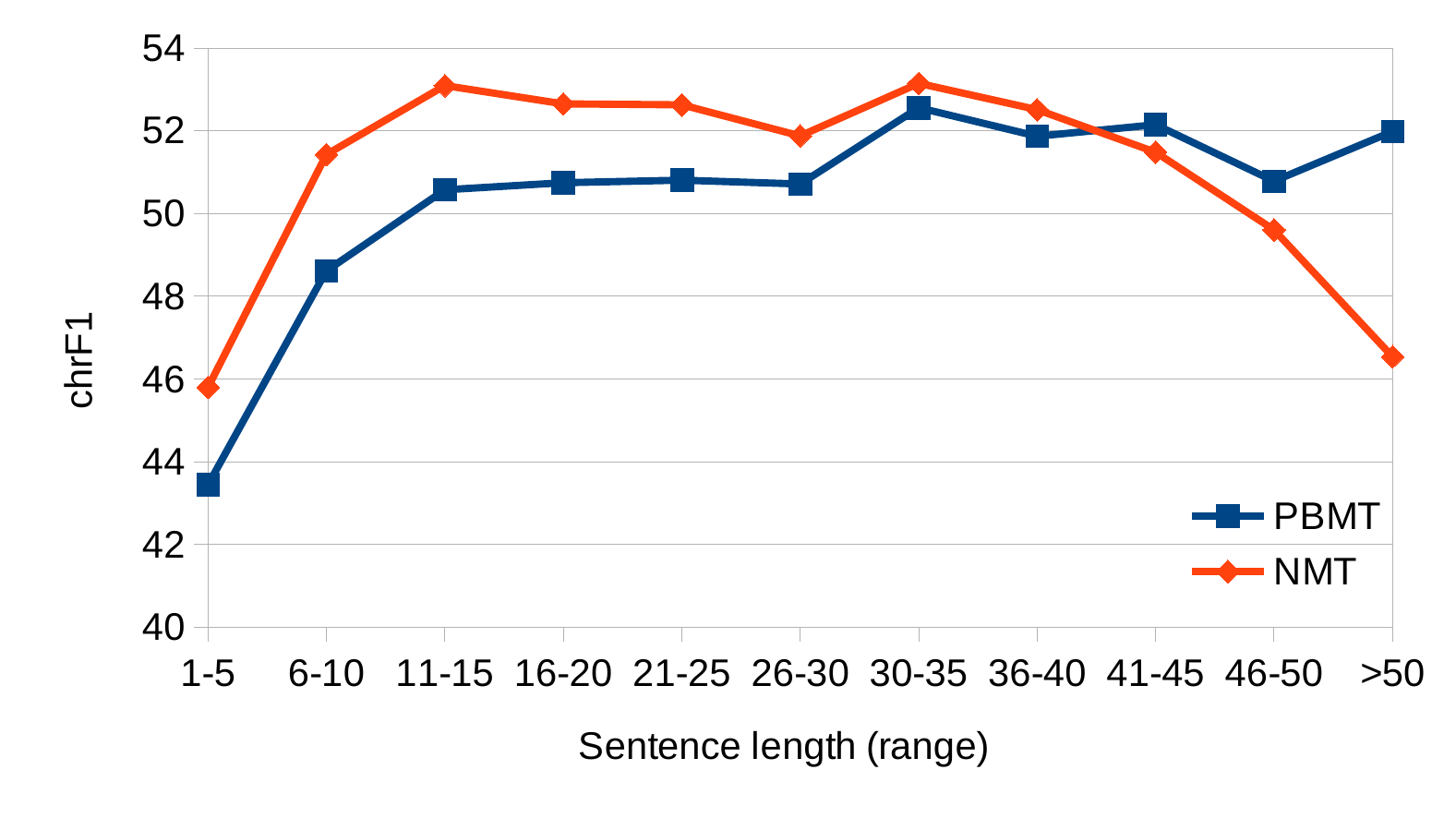}
 \caption{\label{f:length_chrf_enfi}NMT and PBMT chrF1 scores on subsets of different sentence length for \textcolor{purple}{the language direction} EN$\rightarrow$FI.}
\end{figure}

We split the \textcolor{purple}{source side of the} test set in subsets of different lengths:
1 to 5 words (1-5), 6 to 10 and so forth up to 
46 to 50 and finally longer than 50 words ($>50$).
We then evaluate the outputs of the top PBMT and NMT submissions for those subsets with \textcolor{purple}{the chrF1 evaluation metric}.
Figure \ref{f:length_chrf_enfi} presents the results for the language direction EN$\rightarrow$FI.
We can observe that NMT outperforms PBMT up to sentences of length 36-40, while for longer sentences PBMT outperforms NMT, with PBMT's performance remaining fairly stable while NMT's clearly decreases with sentence length.
The results for the other language directions exhibit similar trends.

\begin{figure}[htbp]
 \centering
 \includegraphics[width=0.48\textwidth]{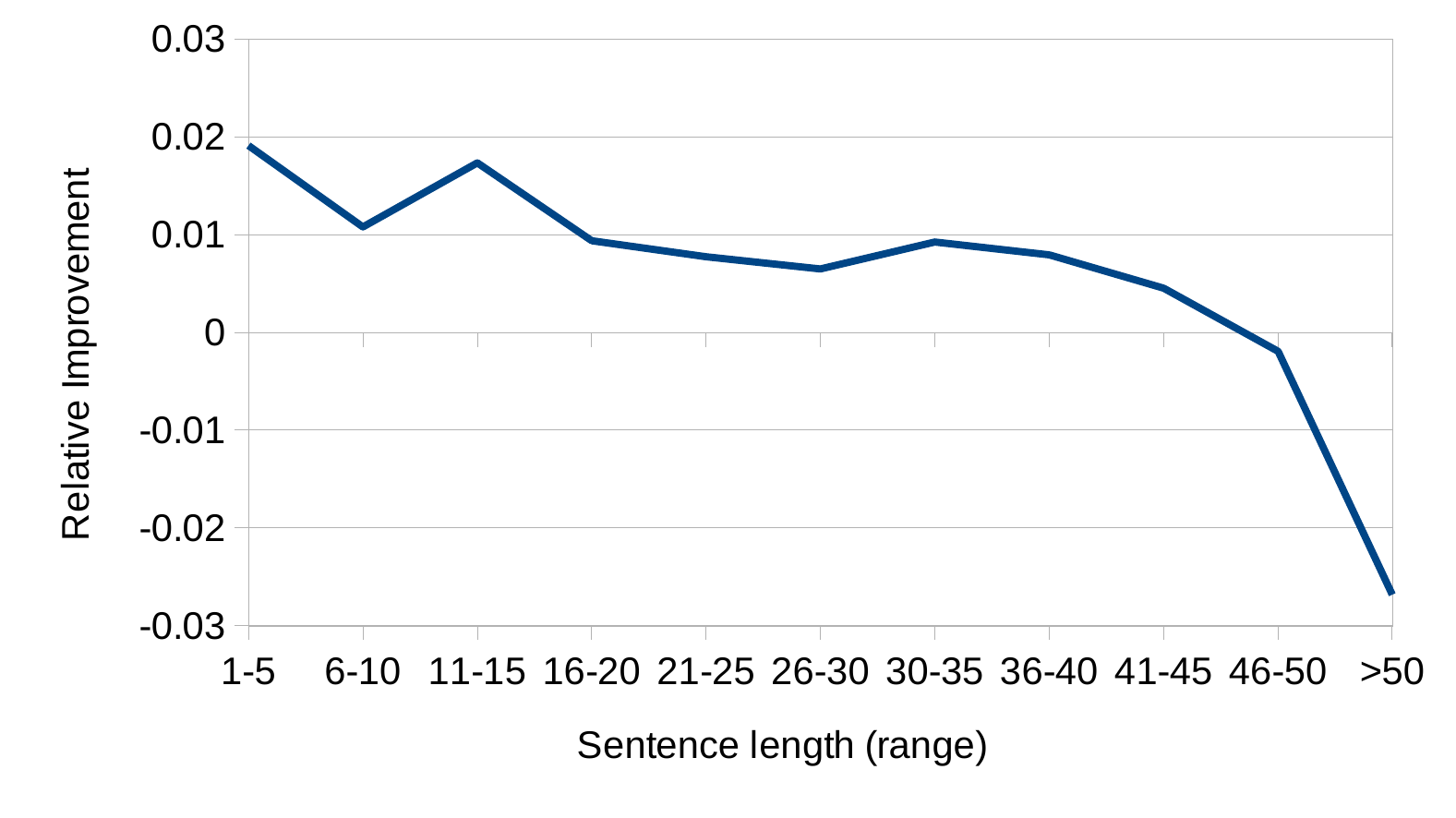}
 \caption{\label{f:length_chrf_avg}Relative improvement of the best NMT versus the best PBMT submission on chrF1 for different sentence lengths, averaged over all the language pairs considered.}
\end{figure}

Figure \ref{f:length_chrf_avg} shows the relative improvements of NMT over PBMT for each sentence length subset, averaged over all the 9 language directions considered.
We observe a clear trend of this value decreasing with sentence length and in fact we found a strong negative Pearson correlation (-0.79) between sentence length and the relative improvement (chrF1) of the best NMT over the best PBMT system.

The correlations for each language direction are shown in Table \ref{t:length_correlations}. 
We observe negative correlations for all the language directions except for DE$\rightarrow$EN.

{\small
\begin{table}[htbp]
\centering
\begin{tabular}{ l | r r r r r}
\bf Direction  & \bf CS	& \bf DE	& \bf FI	& \bf RO	& \bf RU\\
\hline
From EN	&-0.72	&-0.26	&-0.89	&-0.01	&-0.74\\
Into EN	&-0.19	& 0.10	&-		&-0.36	&-0.70\\
\hline
\end{tabular}
\caption{Pearson correlations between sentence length and relative improvement (chrF1) of the best NMT over the best PBMT system for each language pair.}
\label{t:length_correlations}
\end{table}}

\section{Error Categories}\label{s_error_types}

{\small
\begin{table*}
\centering
\begin{tabular}{ l | r r r r r r}
\bf Error type	&\bf EN$\rightarrow$CS&\bf EN$\rightarrow$DE &\bf EN$\rightarrow$FI &\bf EN$\rightarrow$RO &\bf EN$\rightarrow$RU & \bf Average\\
\hline
Inflection	&$-16.18\%$ &$-13.26\%$ &$-11.65\%$ &$-15.13\%$ &$-16.79\%$ &$-14.60\%$\\
Reordering	&$-7.97\%$  &$-21.92\%$ &$-12.12\%$ &$-15.91\%$ &$ -6.18\%$	&$-12.82\%$\\
Lexical		&$-0.44\%$	&$ -3.48\%$	&$ -1.09\%$	&$2.17\%$	&$ -0.09\%$	&$-0.59\%$\\
\hline
\end{tabular}
\caption{Relative improvement of NMT versus PBMT for 3 error categories, for language directions out of English.}
\label{t:hjerson_out_en}
\end{table*}
}

{\small
\begin{table*}
\centering
\begin{tabular}{ l | r r r r r r}
\bf Error type	&\bf CS$\rightarrow$EN&\bf DE$\rightarrow$EN &\bf RO$\rightarrow$EN &\bf RU$\rightarrow$EN & \bf Average\\
\hline
Inflection	&$-4.38\%$ &$-2.47\%$  &$-3.65\%$ &$-21.12\%$ &$-7.91\%$\\
Reordering	&$-8.68\%$ &$-21.09\%$ &$-9.48\%$ &$-8.50\%$  &$-11.94\%$\\
Lexical		&$-1.92\%$ &$-4.91\%$  &$-3.90\%$  &$5.32\%$  &$-1.35\%$\\
\hline
\end{tabular}
\caption{Relative improvement of NMT versus PBMT for 3 error categories, for language directions into English.}
\label{t:hjerson_into_en}
\end{table*}
}

In this experiment we assess the performance of NMT versus PBMT systems on a set of error categories that 
correspond to five word-level error classes: inflection errors,
reordering errors, missing words, extra words and incorrect lexical choices.
These errors are detected automatically using the edit distance, word error rate (WER), precision-based and recall-based position-independent error rates (hPER and rPER, respectively) as implemented in \texttt{Hjerson}~\cite{popovic2011hjerson}.
These error classes are then defined as follows:

\begin{itemize}
\item Inflection error (hINFer). A word for which its full form is marked as a hPER error while its base form matches the \textcolor{blue}{base form in the} reference.
\item Reordering error (hRer). A word that matches the reference but is marked as a WER error.
\item Missing word (MISer). A word that occurs as deletion error in WER, is also a rPER error and does not share the base form with any hypothesis error.
\item Extra word (EXTer). A word that occurs as insertion error in WER, is also a hPER error and does not share the base form with any reference error.
\item Lexical choice error (hLEXer). A word that belongs neither to inflectional errors
nor to missing or extra words.
\end{itemize}

Due to the fact that it is difficult to disambiguate between three of these categories, namely missing words, extra words and lexical choice errors~\cite{Popovic:2011:TAE:2077692.2077694}, we group them in a unique category, which we refer to as lexical errors.

As input, the tool requires the full forms and base forms of the reference translations and MT outputs.
For base forms, 
we use stems for practical reasons.
These are produced with the Snowball stemmer from NLTK\footnote{\url{http://www.nltk.org}} for all languages except for Czech, which is not supported.
For this language we used \textcolor{purple}{the aggresive variant in} \texttt{czech\_stemmer}.\footnote{\url{http://research.variancia.com/czech_stemmer/}}


Tables \ref{t:hjerson_out_en} and \ref{t:hjerson_into_en} show the results for language directions out of English and into English, respectively.
For all language directions, we observe that NMT results in a notable decrease of both inflection ($-14.6\%$ on average for language directions out of EN and $-7.91\%$ for language directions into EN) and reordering ($-12.82\%$ from EN and $-11.94$ into EN) errors.
The reduction of reordering errors is compatible with the
results of the experiment presented in Section~\ref{s_monotonicity}.\footnote{Although the results depicted both in this section and in Section~\ref{s_monotonicity} show that NMT
performs better reordering in general, results for particular language pairs are not exactly the same in both sections. This is due to the fact that the quality of the reordering is computed in different ways. In this section, only those words that match the reference are considered when identifying reordering errors, while in Section~\ref{s_monotonicity} all the words in the sentence are taken into account. That said, in Section~\ref{s_monotonicity} the precision of the results depends on the quality of word alignment.}

Differences in performance for the remaining error category, lexical errors, are much smaller.
In addition, the results for that category show a mixed picture in terms of which paradigm is better, which makes it difficult to derive conclusions that apply regardless of the language pair.
Out of English, NMT results in slightly less errors ($0.59\%$ decrease on average) for all target languages except for RO ($2.17\%$ increase).
Similarly, in the opposite language direction, NMT also results in slightly better performance overall ($1.35\%$ error reduction on average), and looking at individual language directions NMT outperforms PBMT for all of them except RU$\rightarrow$EN.

\section{Conclusions}\label{s_conclusions}

We have conducted a multifaceted evaluation to compare NMT versus PBMT outputs across a number of dimensions for 9 language directions.
Our aim has been to shed more light on the strengths and weaknesses of the newly introduced NMT paradigm, and to check whether, and to what extent, these generalise to different families of source and target languages.
Hereunder we summarise our findings:

\begin{itemize}
\item The outputs of NMT systems are \textcolor{purple}{considerably} different compared to those of PBMT systems. In addition, there is higher inter-system variability in NMT, i.e. outputs by pairs of NMT systems are more different between them than outputs by pairs of PBMT systems.

\item
NMT outputs are more fluent. We have corroborated the results of the manual evaluation of fluency at WMT16, which was conducted only for language directions into English, and we have shown evidence that this finding is true also for directions out of English.

\item 
NMT systems introduce more changes in word order than pure PBMT systems, but less than hierarchical PBMT systems.\footnote{The latter finding applies only to one language direction as only for that one 
the best PBMT system is hierarchical.} Nevertheless, for most language pairs, including those for which the best PBMT system is hierarchical, NMT's reorderings are closer to the reorderings in the reference than those of PBMT.
This corroborates the findings on reordering by Bentivogli et al.~\shortcite{Bentivogli1608.04631}.

\item \textcolor{blue}{
We have found negative correlations between sentence length and the improvement brought by NMT over PBMT for the majority of the languages examined. While for most sentence lengths NMT outperforms PBMT, for very long sentences PBMT outperforms NMT.
The latter was not the case in the work by Bentivogli et al.~\shortcite{Bentivogli1608.04631}.} 
We believe the reason behind this different finding is twofold. Firstly, the average sentence length in their evaluation dataset was considerably shorter; and secondly, the NMT systems included in our evaluation operate on subword units, which increases the effective sentence length they have to deal with.

\item \textcolor{blue}{
NMT performs better in terms of inflection and reordering consistently across all language directions.} We thus confirm that the findings of Bentivogli et al.~\shortcite{Bentivogli1608.04631} regarding these two error types  apply to a wide range of language directions.
\textcolor{purple}{Differences regarding lexical errors are much smaller and inconsistent across language directions; for 7 of them NMT outperforms PBMT while for the remaining 2 the opposite is true.}

\end{itemize}

The results for some of the evaluations, especially error categories (Section ~\ref{s_error_types}) 
have been analysed only superficially, looking at what conclusions can be derived that apply regardless of language direction. Nevertheless, all our data is publicly released,\footnote{\url{https://github.com/antot/neural_vs_phrasebased_smt_eacl17}} so we encourage interested parties to use this resource to conduct deeper language-specific studies.


\section*{Acknowledgments}
The research leading to these results is supported by the European Union Seventh Framework Programme FP7/2007-2013 under grant agreement PIAP-GA-2012-324414 (Abu-MaTran) and by Science Foundation Ireland through the CNGL Programme (Grant 12/CE/I2267) in the ADAPT Centre (www.adaptcentre.ie) at Dublin City University.

\bibliographystyle{eacl2017}
\bibliography{nmt_vs_smt}
\end{document}